\documentclass{article} %

\usepackage[table,dvipsnames,svgnames]{xcolor}

\usepackage{colm2024_conference}

\usepackage{microtype}
\definecolor{citecolor}{HTML}{2779af}
\definecolor{linkcolor}{HTML}{c0392b}
\usepackage[pagebackref=false,breaklinks=true,colorlinks,bookmarks=false,citecolor=citecolor,linkcolor=linkcolor]{hyperref}
\usepackage{url}
\usepackage{booktabs}

\usepackage{tabularx}
\usepackage{multirow}
\usepackage{booktabs}
\usepackage{tabulary}  %
\usepackage[abs]{overpic}
\usepackage[skins]{tcolorbox}
\usepackage{listings}
\usepackage{enumitem}
\usepackage{soul}

\definecolor{codegreen}{rgb}{0,0.6,0}
\definecolor{codegray}{rgb}{0.5,0.5,0.5}
\definecolor{codepurple}{rgb}{0.58,0,0.82}
\definecolor{backcolour}{rgb}{0.95,0.95,0.92}

\lstdefinestyle{mystyle}{
    backgroundcolor=\color{backcolour},
    commentstyle=\color{codegreen},
    keywordstyle=\color{magenta},
    stringstyle=\color{codepurple},
    basicstyle=\ttfamily\footnotesize,
    breakatwhitespace=false,         
    breaklines=true,                 
    keepspaces=true,                
    showspaces=false,                
    showstringspaces=false,
    showtabs=false,                  
    tabsize=2
}

\definecolor{lightgreen}{HTML}{EDF8FB}
\definecolor{mediumgreen}{HTML}{66C2A4}
\definecolor{darkgreen}{HTML}{006D2C}

\definecolor{CMpurple}{rgb}{0.6,0.18,0.64}

\usepackage{caption}
\usepackage{verbatim}

\usepackage{dsfont}
\usepackage{amsmath}
\usepackage{amssymb}
\usepackage{mathtools}
\usepackage{amsthm}

\newcommand{\blockcomment}[1]{}

\DeclareMathOperator*{\argmax}{argmax}

\newcommand{\cD}{\mathcal{D}}

\newcommand{\cP}{\mathcal{P}}
\newcommand{\cQ}{\mathcal{Q}}

\renewcommand{\a}{\alpha}

\title{
Mapping the Increasing Use of LLMs in Scientific Papers
}

\author{Weixin Liang\thanks{Co-first authors, Correspondence to: Weixin Liang \texttt{<wxliang@stanford.edu>}},\, 
Yaohui Zhang\footnotemark[1],\,  Zhengxuan Wu\footnotemark[1],\, Haley Lepp,\\
Stanford University\\
\AND 
Wenlong Ji,\, Xuandong Zhao,\\
Stanford University, UC Santa Barbara\\
\AND 
Hancheng Cao,\, Sheng Liu,\, Siyu He,\, Zhi Huang,\, Diyi Yang, \\
Stanford University\\
\AND
Christopher Potts\footnotemark[2],\, Christopher D Manning\footnotemark[2],\, James Y. Zou\thanks{Co-supervised project, Correspondence to: James Zou \texttt{<jamesz@stanford.edu>}} \\
Stanford University\\
}

\colmfinalcopy %
\begin{document}

\maketitle

\begin{abstract}

Scientific publishing lays the foundation of science by disseminating research findings, fostering collaboration, encouraging reproducibility, and ensuring that scientific knowledge is accessible, verifiable, and built upon over time. Recently, there has been immense speculation about how many people are using large language models (LLMs) like ChatGPT in their academic writing, and to what extent this tool might have an effect on global scientific practices. However, we lack a precise measure of the proportion of academic writing substantially modified or produced by LLMs. 
To address this gap, we conduct the first systematic, large-scale analysis across 950,965 papers published between January 2020 and February 2024 on the \textit{arXiv}, \textit{bioRxiv}, and \textit{Nature} portfolio journals, using a population-level statistical framework to measure the prevalence of LLM-modified content over time. Our statistical estimation operates on the corpus level and is more robust than inference on  individual instances. Our findings reveal a steady increase in LLM usage, with the largest and fastest growth observed in Computer Science papers (up to 17.5\%). In comparison, Mathematics papers and the Nature portfolio showed the least LLM modification (up to 6.3\%). Moreover, at an aggregate level, our analysis reveals that higher levels of LLM-modification are associated with papers whose first authors post preprints more frequently, papers in more crowded research areas, and papers of shorter lengths. Our findings suggests that LLMs are being broadly used in scientific writings.

\end{abstract}

\begin{figure}[ht!] 
    \centering
    \includegraphics[width=1.00\textwidth]{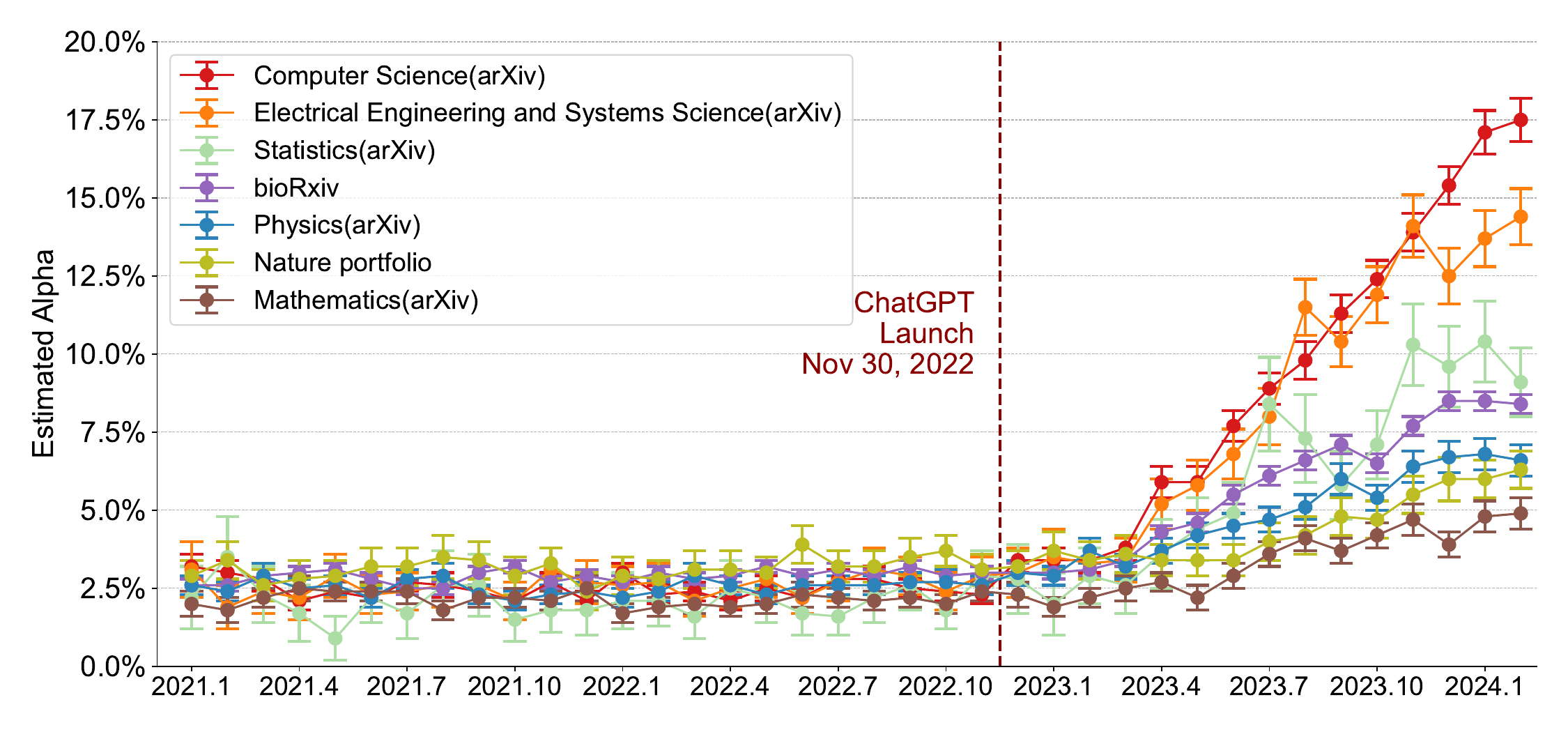}
    \caption{
    \textbf{Estimated Fraction of LLM-Modified Sentences across Academic Writing Venues over Time. }
    This figure displays the fraction ($\a$) of sentences estimated to have been substantially modified by LLM in abstracts from various academic writing venues. The analysis includes five areas within \textit{arXiv} (Computer Science, Electrical Engineering and Systems Science, Mathematics, Physics, Statistics), articles from \textit{bioRxiv}, and a combined dataset from 15 journals within the \textit{Nature} portfolio. Estimates are based on the \textit{distributional GPT quantification} framework, 
    which provides population-level estimates rather than individual document analysis. Each point in time is independently estimated, with no temporal smoothing or continuity assumptions applied.  
    Error bars indicate 95\% confidence intervals by bootstrap.
    Further analysis of paper introductions is presented in Figure~\ref{fig: temporal-introduction}.
    }
    \label{fig: temporal-abstract}
\end{figure}

\begin{figure}[ht!]
    \centering
    \includegraphics[width=1.00\textwidth]{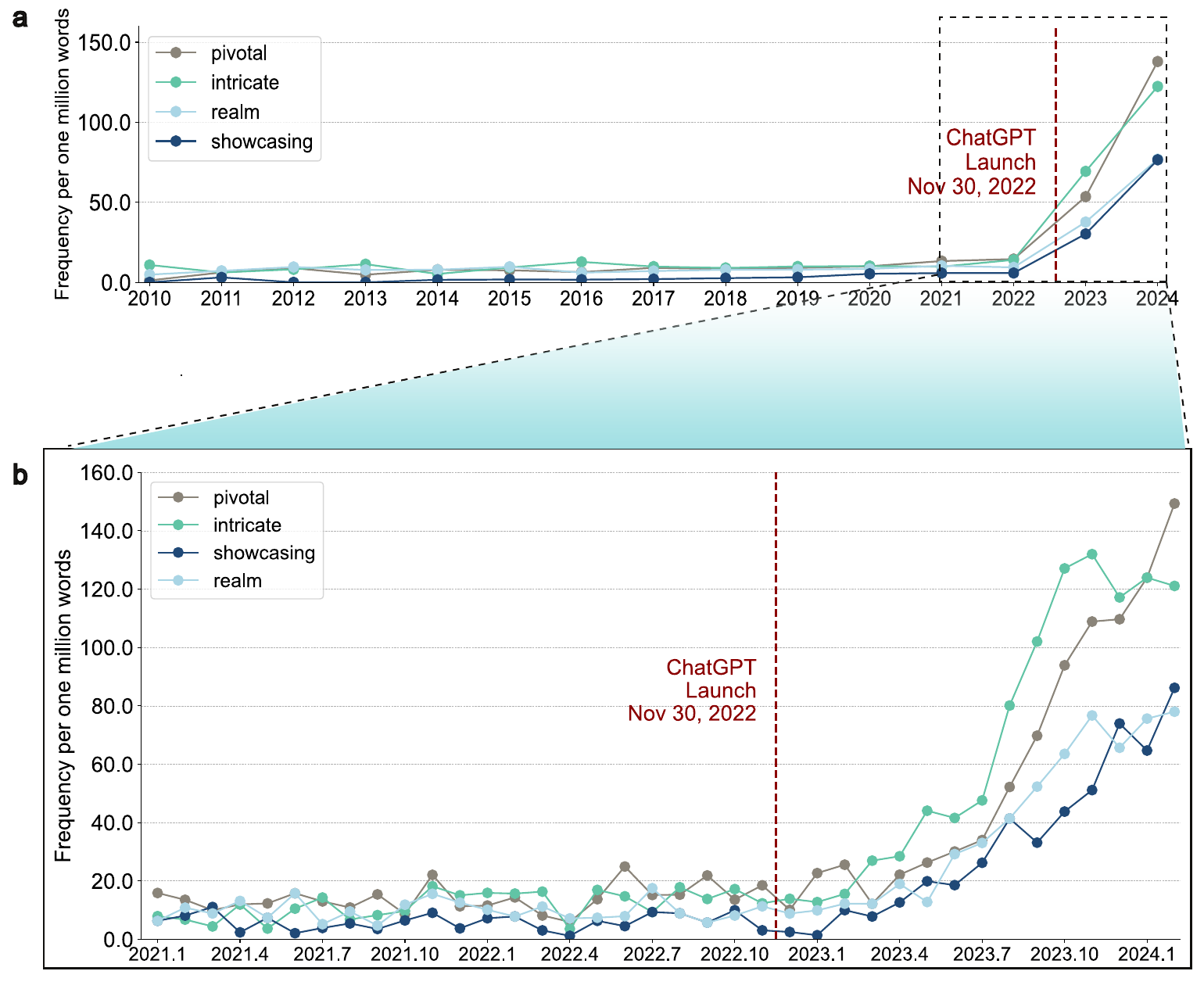}
\caption{
    \textbf{Word Frequency Shift in arXiv Computer Science abstracts over 14 years (2010-2024).} 
The plot shows the frequency over time for the top 4 words most disproportionately used by LLM compared to humans, as measured by the log odds ratio. The words are: \textit{realm}, \textit{intricate}, \textit{showcasing}, \textit{pivotal}. 
These terms maintained a consistently low frequency in arXiv CS abstracts over more than a decade (2010--2022) but experienced a sudden surge in usage starting in 2023.
}
\label{fig: arxiv-revisions}
\end{figure}

\section{Introduction}
\label{sec:introduction}

Since the release of ChatGPT in late 2022, anecdotal examples of both published papers \citep{Okunyte2023GoogleSearch, Deguerin24} and peer reviews \citep{Oransky24} which appear to be ChatGPT-generated have inspired humor and concern.\footnote{Increased attention to ChatGPT-use by multilingual scholars has also brought to the fore important conversations about entrenched linguistic discrimination in academic publishing \citep{khanna22}.} While certain tells, such as ``regenerate response” \citep{Conroy2023Nature, Conroy20232Nature} and ``as an AI language model" \citep{Vincent2023AITextPollution}, found in published papers indicate modified content, less obvious cases are nearly impossible to detect at the individual level \citep{Nature-news-Abstracts-written-by-ChatGPT-fool-scientists,Abstracts-written-by-ChatGPT-fool-scientists}. \cite{liang2024monitoring} present a method for detecting the percentage of LLM-modified text in a corpus beyond such obvious cases. Applied to scientific publishing, the importance of this at-scale approach is two-fold: first, rather than looking at LLM-use as a type of rule-breaking on an individual level, we can begin to uncover structural circumstances which might motivate its use. Second, by examining LLM-use in academic publishing at-scale, we can capture epistemic and linguistic shifts, miniscule at the individual level, which become apparent with a birdseye view.

Measuring the extent of LLM-use on scientific publishing has urgent applications. Concerns about accuracy, plagiarism, anonymity, and ownership have prompted some prominent scientific institutions to take a stance on the use of LLM-modified content in academic publications. The International Conference on Machine Learning (ICML) 2023, a major machine learning conference, has prohibited the inclusion of text generated by LLMs like ChatGPT in submitted papers, unless the generated text is used as part of the paper's experimental analysis~\citep{ICML2023LLMPolicy}. Similarly, the journal \textit{Science} has announced an update to their editorial policies, specifying that text, figures, images, or graphics generated by ChatGPT or any other LLM tools cannot be used in published works~\citep{doi:10.1126/science.adg7879}. Taking steps to measure the extent of LLM-use can offer a first-step in identifying risks to the scientific publishing ecosystem. 
Furthermore, exploring the circumstances in which LLM-use is high can offer publishers and academic institutions useful insight into author behavior. Sites of high LLM-use can act as indicators for structural challenges faced by scholars. These range from pressures to ``publish or perish'' which encourage rapid production of papers to concerns about linguistic discrimination that might lead authors to use LLMs as prose editors.

We conduct the first systematic, large-scale analysis to quantify the prevalence of LLM-modified content across multiple academic platforms, extending a recently proposed, state-of-the-art \textit{distributional GPT quantification} framework~\citep{liang2024monitoring} for estimating the fraction of AI-modified content in a corpus. Throughout this paper, we use the term ``LLM-modified'' to refer to text content substantially updated by ChatGPT beyond basic spelling and grammatical edits. Modifications we capture in our analysis could include, for example, summaries of existing writing or the generation of prose based on outlines.

A key characteristic of this framework is that it operates on the population level, without the need to perform inference on any individual instance. As validated in the prior paper, the framework is orders of magnitude more computationally efficient and thus scalable, produces more accurate estimates, and generalizes better than its counterparts under significant temporal distribution shifts and other realistic distribution shifts. 

We apply this framework to the abstracts and introductions (Figures~\ref{fig: temporal-abstract} and \ref{fig: temporal-introduction}) of academic papers across multiple academic disciplines,including \textit{arXiv}, \textit{bioRxiv}, and 15 journals within the Nature portfolio, such as \textit{Nature}, \
\textit{Nature Biomedical Engineering}, \textit{Nature Human Behaviour}, and \textit{Nature Communications}. Our study analyzes a total of 950,965 papers published between January 2020 and February 2024, comprising 773,147 papers from \textit{arXiv}, 161,280 from \textit{bioRxiv}, and 16,538 from the Nature portfolio journals. The papers from \textit{arXiv} cover multiple academic fields, including Computer Science, Electrical Engineering and Systems Science, Mathematics, Physics, and Statistics. These datasets allow us to quantify the prevalence of LLM-modified academic writing over time and across a broad range of academic fields.

Our results indicate that the largest and fastest growth was observed in Computer Science papers, with $\alpha$ reaching 17.5\% for abstracts and 15.3\% for introductions by February 2024. In contrast, Mathematics papers and the \textit{Nature} portfolio showed the least increase, with $\alpha$ reaching 4.9\% and 6.3\% for abstracts and 3.5\% and 6.4\% for introductions, respectively.
Moreover, our analysis reveals at an aggregate level that higher levels of LLM-modification are associated with papers whose first authors post preprints more frequently and papers with shorter lengths. Results also demonstrate a closer relationship between papers with LLM-modifications, which could indicate higher use in more crowded fields of study (as measured by the distance to the nearest neighboring paper in the embedding space), or that generated-text is flattening writing diversity.

\section{Related Work}

\paragraph{GPT Detectors} 
Various methods have been proposed for detecting LLM-modified text, including zero-shot approaches that rely on statistical signatures characteristic of machine-generated content \citep{Lavergne2008DetectingFC,Badaskar2008IdentifyingRO,Beresneva2016ComputerGeneratedTD,solaiman2019release,Mitchell2023DetectGPTZM,Yang2023DNAGPTDN,Bao2023FastDetectGPTEZ,Tulchinskii2023IntrinsicDE} and training-based methods that finetune language models for binary classification of human vs. LLM-modified text \citep{Bhagat2013SquibsWI,Zellers2019DefendingAN,Bakhtin2019RealOF,Uchendu2020AuthorshipAF,Chen2023GPTSentinelDH,Yu2023GPTPT,Li2023DeepfakeTD,Liu2022CoCoCM,Bhattacharjee2023ConDACD,Hu2023RADARRA}. However, these approaches face challenges such as the need for access to LLM internals, overfitting to training data and language models, vulnerability to adversarial attacks \citep{Wolff2020AttackingNT}, and bias against non-dominant language varieties \citep{Liang2023GPTDA}. The effectiveness and reliability of publicly available LLM-modified text detectors have also been questioned \citep{OpenAIGPT2,jawahar2020automatic,fagni2021tweepfake,ippolito2019automatic,mitchell2023detectgpt,human-hard-to-detect-generated-text,mit-technology-review-how-to-spot-ai-generated-text,survey-2023,solaiman2019release,Kirchner2023,Kelly2023}, with the theoretical possibility of accurate instance-level detection being debated \citep{Weber-Wulff2023,Sadasivan2023CanAT,chakraborty2023possibilities}. In this study, we apply the recently proposed \textit{distributional GPT quantification} framework \citep{liang2024monitoring}, which estimates the fraction of LLM-modified content in a text corpus at the population level, circumventing the need for classifying individual documents or sentences and improving upon the stability, accuracy, and computational efficiency of existing approaches. 
A more comprehensive discussion of related work can be found in Appendix~\ref{appendix:sec:related-work}.

\section{Background: the \textit{distributional LLM quantification} framework} \label{main:sec: method}

We adapt the \textit{distributional LLM quantification} framework from \cite{liang2024monitoring} to quantify the use of AI-modified academic writing. The framework consists of the following steps:

\begin{enumerate}[nosep, leftmargin=2em] %
  \item \textbf{Problem formulation}: Let $\cP$ and $\cQ$ be the probability distributions of human-written and LLM-modified documents, respectively. The mixture distribution is given by $\cD_\alpha(X) = (1-\alpha)\cP(x)+\alpha \cQ(x)$, where $\alpha$ is the fraction of AI-modified documents. The goal is to estimate $\alpha$ based on observed documents $\{X_i\}_{i=1}^N \sim \cD_{\alpha}$.

  \item \textbf{Parameterization}: To make $\alpha$ identifiable, the framework models the distributions of token occurrences in human-written and LLM-modified documents, denoted as $\cP_T$ and $\cQ_T$, respectively, for a chosen list of tokens $T=\{t_i\}_{i=1}^M$. The occurrence probabilities of each token in human-written and LLM-modified documents, $p_t$ and $q_t$, are used to parameterize $\cP_T$ and $\cQ_T$:
  \begin{align*}
    \cP_T(X) = \prod_{t\in T}p_t^{\mathbb{1}\{t\in X\}}(1-p_t)^{\mathbb{1}\{t\notin X\}}, \quad \cQ_T(X) = \prod_{t\in T}q_t^{\mathbb{1}\{t\in X\}}(1-q_t)^{\mathbb{1}\{t\notin X\}}.
  \end{align*}

  \item \textbf{Estimation}: The occurrence probabilities $p_t$ and $q_t$ are estimated using collections of known human-written and LLM-modified documents, $\{X_j^P\}_{j=1}^{n_P}$ and $\{X_j^Q\}_{j=1}^{n_Q}$, respectively:
  \begin{align*}
    \hat p_t = \frac{1}{n_P}\sum_{j=1}^{n_P}\mathbb{1}\{t\in X_j^P\}, \quad \hat q_t = \frac{1}{n_Q}\sum_{j=1}^{n_Q}\mathbb{1}\{t\in X_j^Q\}.
  \end{align*}

  \item \textbf{Inference}: The fraction $\alpha$ is estimated by maximizing the log-likelihood of the observed documents under the mixture distribution $\hat\cD_{\alpha, T}(X) = (1-\alpha)\hat\cP_{T}(X)+\alpha \hat\cQ_{T}(X)$:
  \begin{align*}
    \hat\alpha^{\text{MLE}}_T = \argmax_{\alpha\in [0,1]}\sum_{i=1}^N\log\left((1-\alpha)\hat\cP_{T}(X_i)+\alpha \hat\cQ_{T}(X_i)\right).
  \end{align*}
\end{enumerate}

\cite{liang2024monitoring} demonstrate that the data points $\{X_i\}_{i=1}^N \sim \cD_{\alpha}$ can be constructed either as a document or as a sentence, and both work well. Following their method, we use sentences as the unit of data points for the estimates for the main results. In addition, we extend this framework for our application to academic papers with two key differences:

\paragraph{Generating Realistic LLM-Produced Training Data}

We use a two-stage approach to generate LLM-produced text, as simply prompting an LLM with paper titles or keywords would result in unrealistic scientific writing samples containing fabricated results, evidence, and ungrounded or hallucinated claims.

Specifically, given a paragraph from a paper known to not include LLM-modification, we first perform abstractive summarization using an LLM to extract key contents in the form of an outline. We then prompt the LLM to generate a full paragraph based the outline (see Appendix for full prompts).

Our two-stage approach can be considered a \textit{counterfactual} framework for generating LLM text: \textit{given a paragraph written entirely by a human, how would the text read if it conveyed almost the same content but was generated by an LLM?} This additional abstractive summarization step can be seen as the control for the content. 
This approach also simulates how scientists may be using LLMs in the writing process, where the scientists first write the outline themselves and then use LLMs to generate the full paragraph based on the outline.

\begin{figure}[htb]
\centering
\includegraphics[width=1.0\textwidth]{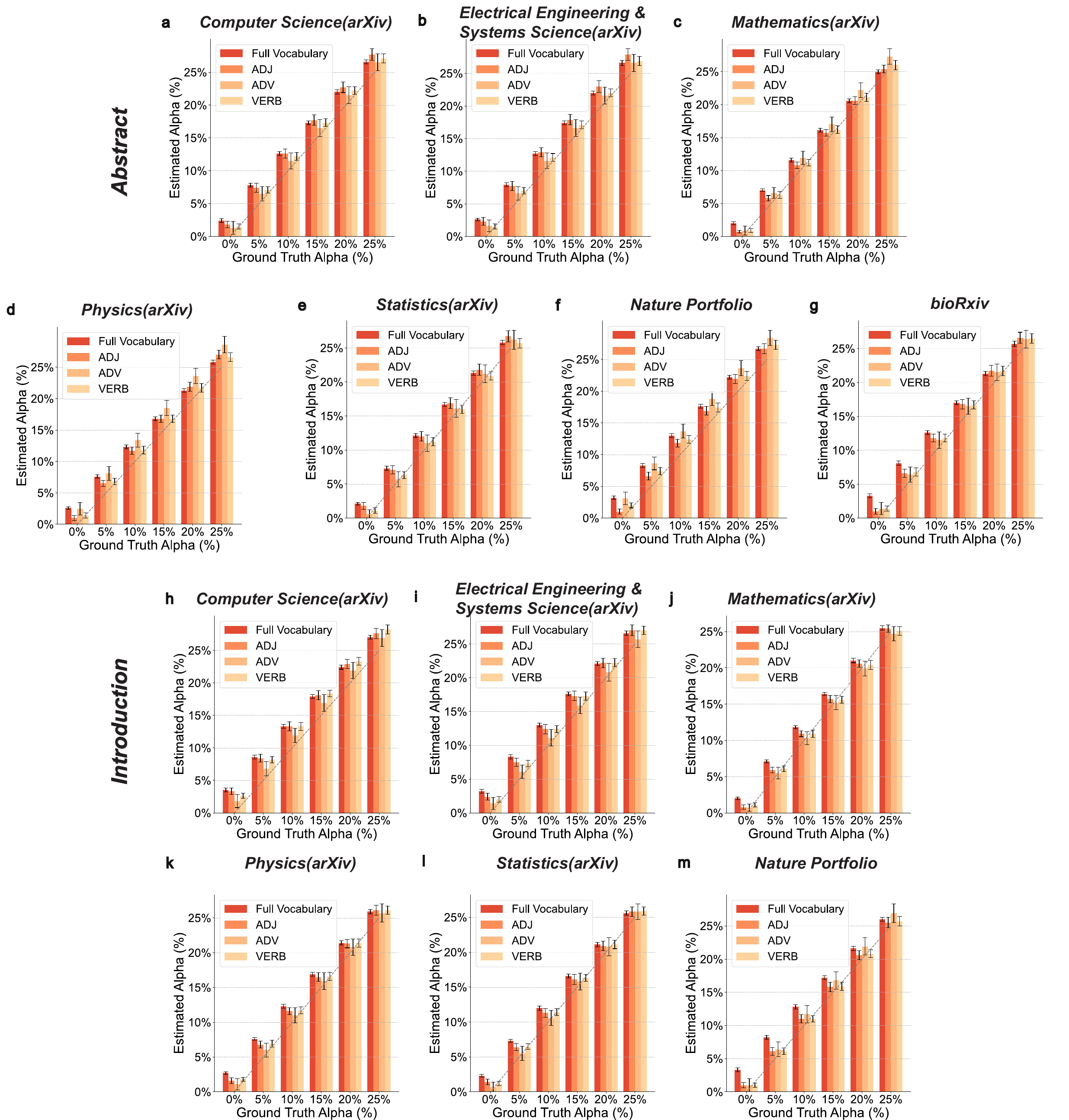}
\caption{
\textbf{Fine-grained Validation of Model Performance Under Temporal Distribution Shift.}
We evaluate the accuracy of our models in estimating the fraction of LLM-modified content ($\alpha$) under a challenging temporal data split, where the validation data (sampled from 2022-01-01 to 2022-11-29) are temporally separated from the training data (collected up to 2020-12-31) by at least a year.
The X-axis indicates the ground truth $\alpha$, while the Y-axis indicates the model's estimated $\alpha$. In all cases, the estimation error for $\alpha$ is less than 3.5\%. 
The first 7 panels (a--g) are the validation on abstracts for each academic writing venue, while the later 6 panels (h--m) are the validation on introductions. We did not include \textit{bioRxiv} introductions due to the unavailability of bulk PDF downloads.  Error bars indicate 95\% confidence intervals by bootstrap.
}
\label{fig: validations}
\end{figure}

\paragraph{Using the Full Vocabulary for Estimation}
We use the full vocabulary instead of only adjectives, as our validation shows that adjectives, adverbs, and verbs all perform well in our application (Figure~\ref{fig: validations}). 
Using the full vocabulary minimizes design biases stemming from vocabulary selection. We also find that using the full vocabulary is more sample-efficient in producing stable estimates, as indicated by their smaller confidence intervals by bootstrap.

\section{Implementation and Validations}
\label{main:sec:implementation-and-validations}

\subsection{Data Collection and Sampling}
\label{main:subsec:data}

We collect data from three sources: \textit{arXiv}, \textit{bioRxiv}, and 15 journals from the \textit{Nature} portfolio. For each source, we randomly sample up to 2,000 papers per month from January 2020 to February 2024. The procedure for generating the LLM-generated corpus data is described in Section $\S$~\ref{main:sec: method}. We focused on the introduction sections for the main texts, as the introduction was the most consistently and commonly occurring section across diverse categories of papers. 
See Appendix~\ref{appendix:sec:implementation} for comprehensive implementation details.

\subsection{Data Split, Model Fitting, and Evaluation}
\label{main:subsec:training-validation}

For model training, we count word frequencies for scientific papers written before the release of ChatGPT and the LLM-modified corpora described in Section \ref{main:sec: method}. We fit the model with data from 2020, and use data from January 2021 onwards for validation and inference. We fit separate models for abstracts and introductions for each major category. 

To evaluate model accuracy and calibration under temporal distribution shift, we use 3,000 papers from January 1, 2022, to November 29, 2022, a time period prior to the release of ChatGPT, as the validation data. We construct validation sets with LLM-modified content proportions ($\alpha$) ranging from 0\% to 25\%, in 5\% increments, and compared the model's estimated $\alpha$ with the ground truth $\alpha$ (Figure \ref{fig: validations}). Full vocabulary, adjectives, adverbs, and verbs all performed well in our application, with a prediction error consistently less than 3.5\% at the population level across various ground truth $\alpha$ values (Figure \ref{fig: validations}).

\section{Main Results and Findings}
\label{sec:Results}

\subsection{Temporal Trends in AI-Modified Academic Writing}
\label{subsec:main-results}

\paragraph{Setup}
We apply the model to estimate the fraction of LLM-modified content ($\alpha$) for each paper category each month, for both abstracts and introductions. Each point in time was independently estimated, with no temporal smoothing or continuity assumptions applied.

\paragraph{Results}
Our findings reveal a steady increase in the fraction of AI-modified content ($\alpha$) in both the abstracts (Figure~\ref{fig: temporal-abstract}) and the introductions (Figure~\ref{fig: temporal-introduction}), with the largest and fastest growth observed in Computer Science papers. By February 2024, the estimated $\alpha$ for Computer Science had increased to 17.5\% for abstracts and 15.5\% for introductions. The second-fastest growth was observed in Electrical Engineering and Systems Science, with the estimated $\alpha$ reaching 14.4\% for abstracts and 12.4\% for introductions during the same period. In contrast, Mathematics papers and the \textit{Nature} portfolio showed the least increase. By the end of the studied period, the estimated $\alpha$ for Mathematics had increased to 4.9\% for abstracts and 3.9\% for introductions, while the estimated $\alpha$ for the Nature portfolio had reached 6.3\% for abstracts and 4.3\% for introductions.

The November 2022 estimates serve as a pre-ChatGPT reference point for comparison, as ChatGPT was launched on November 30, 2022. The estimated $\alpha$ for Computer Science in November 2022 was 2.3\%, while for Electrical Engineering and Systems Science, Mathematics, and the Nature portfolio, the estimates were 2.9\%, 2.4\%, and 3.1\%, respectively. These values are consistent with the false positive rate reported in the earlier section ($\S$~\ref{main:subsec:training-validation}).

\subsection{Relationship Between First-Author Preprint Posting Frequency and GPT Usage}
\label{subsec:preprint-frequency}

\begin{figure}[htb] 
    \centering
    \includegraphics[width=1.00\textwidth]{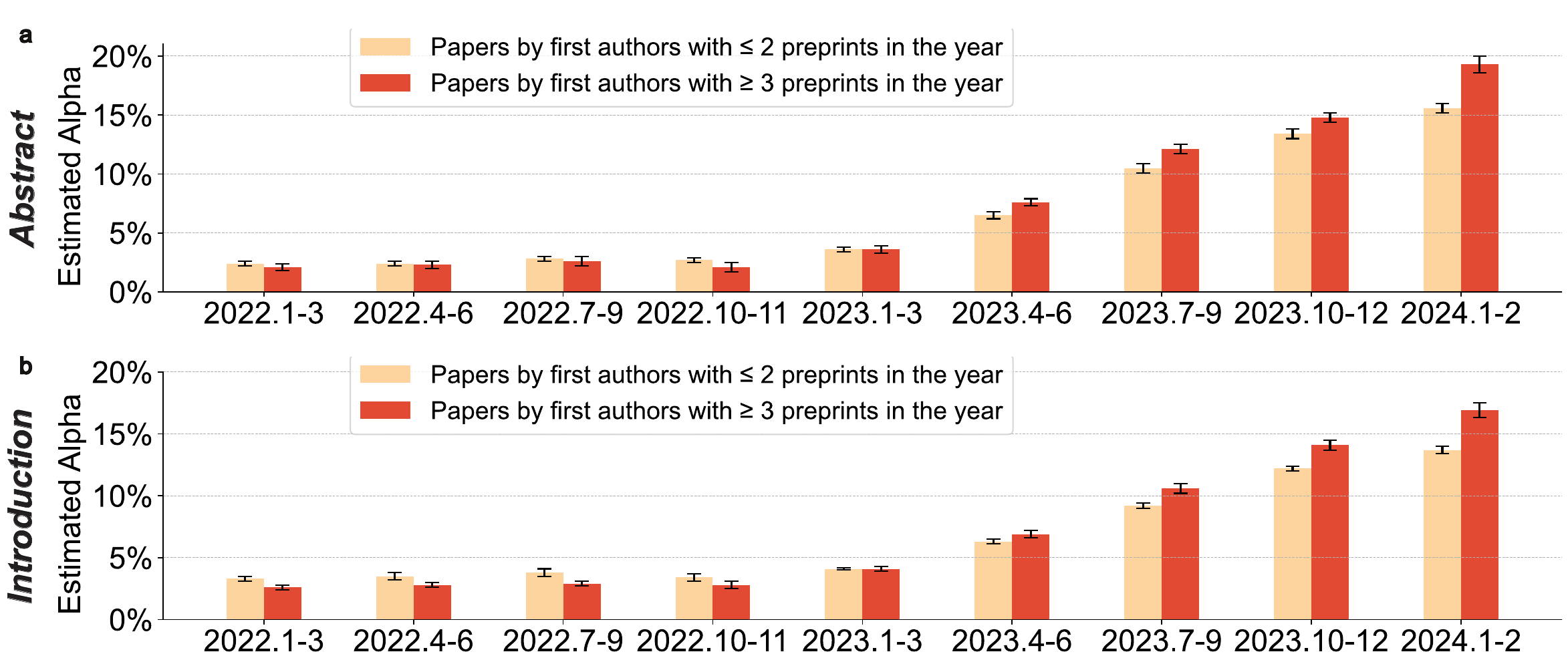}
    \caption{    
\textbf{Papers authored by first authors who post preprints more frequently tend to have a higher fraction of LLM-modified content.}
Papers in \textit{arXiv} Computer Science are stratified into two groups based on the preprint posting frequency of their first author, as measured by the number of first-authored preprints in the year.  Error bars indicate 95\% confidence intervals by bootstrap.
    }
    \label{fig: preprint-frequency}
\end{figure}

We found a notable correlation between the number of preprints posted by the first author on \textit{arXiv} and the estimated number of LLM-modified sentences in their academic writing. Papers were stratified into two groups based on the number of first-authored \textit{arXiv} Computer Science preprints by the first author in the year: those with two or fewer ($\leq 2$) preprints and those with three or more ($\geq 3$) preprints (Figure \ref{fig: preprint-frequency}). We used the 2023 author grouping for the 2024.1-2 data, as we don't have the complete 2024 author data yet.

By February 2024, abstracts of papers whose first authors had $\geq 3$ preprints in 2023 showed an estimated 19.3\% of sentences modified by AI, compared to 15.6\% for papers whose first authors had $\leq 2$ preprints (Figure \ref{fig: preprint-frequency}a). We observe a similar trend in the introduction sections, with first authors posting more preprints having an estimated 16.9\% LLM-modified sentences, compared to 13.7\% for first authors posting fewer preprints (Figure \ref{fig: preprint-frequency}b).
Since the first-author preprint posting frequency may be confounded by research field, we conduct an additional robustness check for our findings. We find that the observed trend holds for each of the three \textit{arXiv} Computer Science sub-categories: cs.CV (Computer Vision and Pattern Recognition), cs.LG (Machine Learning), and cs.CL (Computation and Language) (Supp Figure~\ref{supp:figure:upload}).

Our results suggest that researchers posting more preprints tend to utilize LLMs more extensively in their writing. One interpretation of this effect could be that the increasingly competitive and fast-paced nature of CS research communities incentivizes taking steps to accelerate the writing process. We do not evaluate whether these preprints were accepted for publication.

\subsection{Relationship Between Paper Similarity and LLM Usage} 
\label{subsec:crowdedness}

We investigate the relationship between a paper's similarity to its closest peer and the estimated LLM usage in the abstract. To measure similarity, we first embed each abstract from the \textit{arXiv} Computer Science papers using OpenAI's text-embedding-ada-002 model, creating a vector representation for each abstract. We then calculate the distance between each paper's vector and its nearest neighbor within the \textit{arXiv} Computer Science abstracts. Based on this similarity measure we divide papers into two groups: those more similar to their closest peer (below median distance) and those less similar (above median distance).

\begin{figure}[htb]
\centering
\includegraphics[width=1.00\textwidth]{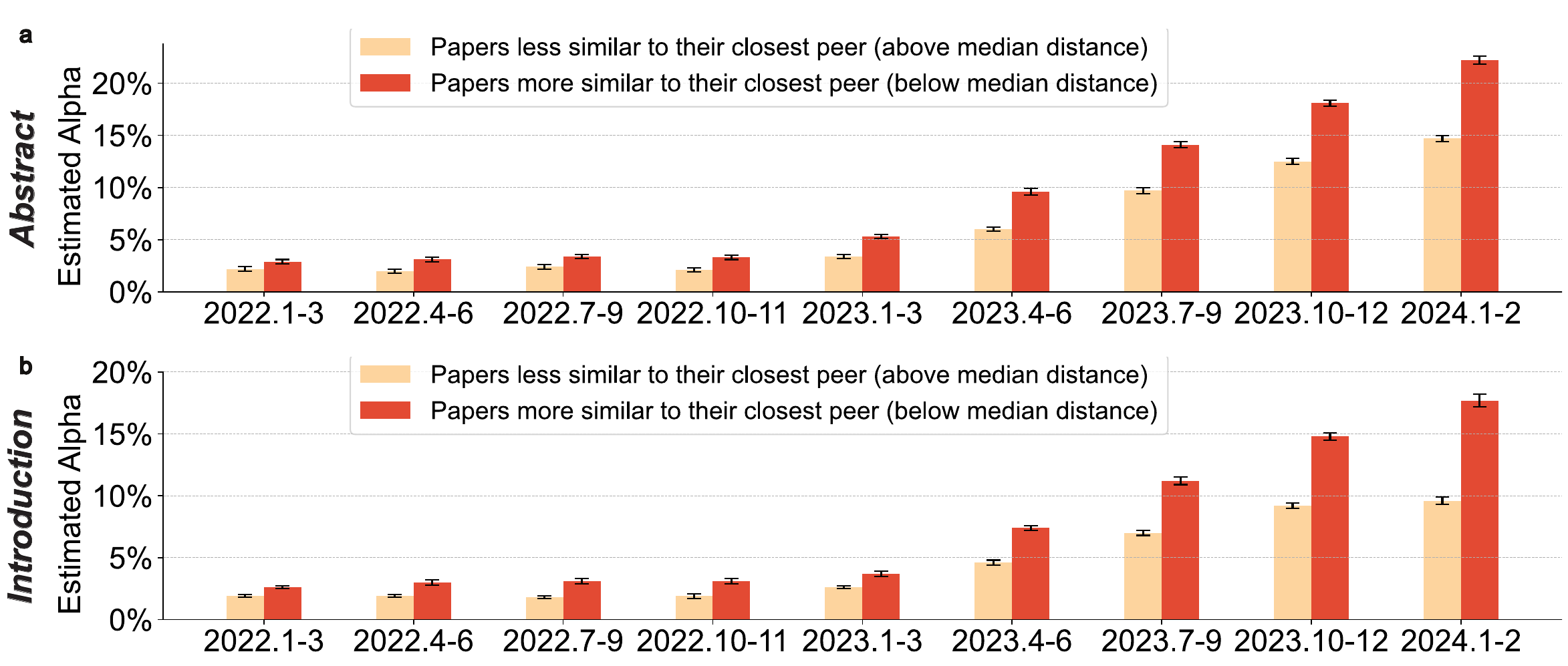}
\caption{
\textbf{Papers in more crowded research areas tend to have a higher fraction of LLM-modified content.}
Papers in \textit{arXiv} Computer Science are divided into two groups based on their abstract's embedding distance to their closest peer: papers more similar to their closest peer (below median distance) and papers less similar to their closest peer (above median distance). Error bars indicate 95\% confidence intervals by bootstrap.
}
\label{fig: homogenization}
\end{figure}

The temporal trends of LLM usage for these two groups are shown in Figure \ref{fig: homogenization}. After the release of ChatGPT, papers most similar to their closest peer consistently showed higher LLM usage compared to those least similar. By February 2024, the abstracts of papers more similar to their closest peer had an estimated 22.2\% of sentences modified by LLMs, compared to 14.7\% for papers less similar to their closest peer.
To account for potential confounding effects of research fields, we conducted an additional robustness check by measuring the nearest neighbor distance within each of the three \textit{arXiv} Computer Science sub-categories: cs.CV (Computer Vision and Pattern Recognition), cs.LG (Machine Learning), and cs.CL (Computation and Language), and found that the observed trend holds for each sub-category (Supp Figure~\ref{supp:figure:crowded}).

There are several ways to interpret these findings. First, LLM-use in writing could cause the similarity in writing or content. Community pressures may even motivate scholars to try to sound more similar -- to assimilate to the ``style" of text generated by an LLM.
Alternatively,  LLMs may be more commonly used in research areas where papers tend to be more similar to each other. This could be due to the competitive nature of these crowded subfields, which may pressure researchers to write faster and produce similar findings. Future interdisciplinary research should explore these hypotheses.

\subsection{Relationship Between Paper Length and AI Usage}
\label{subsec:length}

We also explored the association between paper length and LLM usage in \textit{arXiv} Computer Science papers. Papers were stratified by their full text word count, including appendices, into two bins: below or above 5,000 words (the rounded median).

\begin{figure}[htb]
\centering
\includegraphics[width=1.00\textwidth]{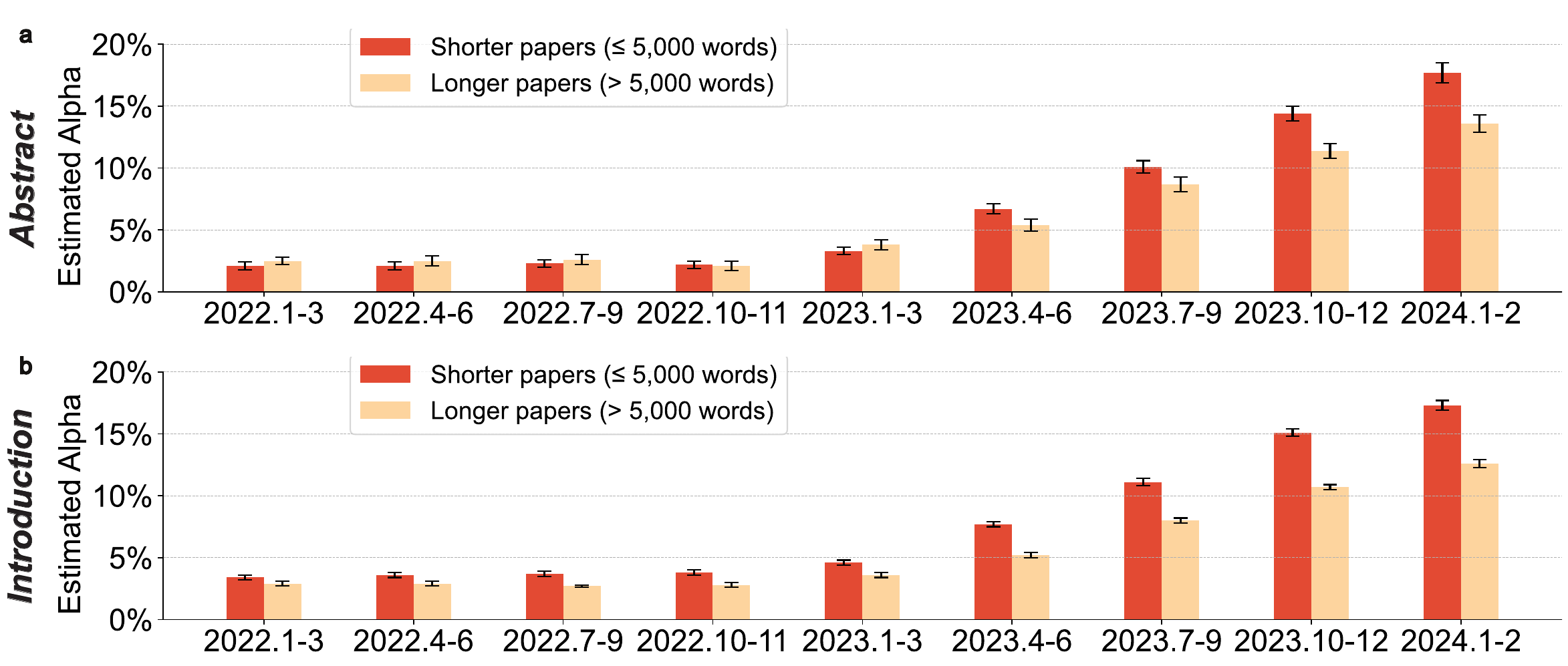}
\caption{
\textbf{Shorter papers tend to have a higher fraction of LLM-modified content.}
\textit{arXiv} Computer Science papers are stratified by their full text word count, including appendices, into two bins: below or above 5,000 words (the rounded median). 
Error bars indicate 95\% confidence intervals by bootstrap.
}
\label{fig: temporal-full-length}
\end{figure}

Figure \ref{fig: temporal-full-length} shows the temporal trends of LLM usage for these two groups. After the release of ChatGPT, shorter papers consistently showed higher AI usage compared to longer papers. By February 2024, the abstracts of shorter papers had an estimated 17.7\% of sentences modified by LLMs, compared to 13.6\% for longer papers (Figure \ref{fig: temporal-full-length}a). A similar trend was observed in the introduction sections (Figure \ref{fig: temporal-full-length}b). 
To account for potential confounding effects of research fields, we conducted an additional robustness check. The finding holds for both cs.CV (Computer Vision and Pattern Recognition) and cs.LG (Machine Learning) (Supp Figure~\ref{supp:figure:length}). However, for cs.CL (Computation and Language), we found no significant difference in LLM usage between shorter and longer papers, possibly due to the limited sample size, as we only parsed a subset of the PDFs and calculated their full length.

As Computer Science conference papers typically have a fixed page limit, longer papers likely have more substantial content in the appendix. The lower LLM usage in these papers may suggest that researchers with more comprehensive work rely less on LLM-assistance in their writing. However, further investigation is needed to determine the relationship between paper length, content comprehensiveness, and the quality of the research.

\section{Discussion}

Our findings show a sharp increase in the estimated fraction of LLM-modified content in academic writing beginning about five months after the release of ChatGPT, with the fastest growth observed in Computer Science papers. This trend may be partially explained by Computer Science researchers' familiarity with and access to large language models. Additionally, the fast-paced nature of LLM research and the associated pressure to publish quickly may incentivize the use of LLM writing assistance \citep{foster2015tradition}.

We expose several other factors associated with higher LLM usage in academic writing. First, authors who post preprints more frequently show a higher fraction of LLM-modified content in their writing. Second, papers in more crowded research areas, where papers tend to be more similar, showed higher LLM-modification compared to those in less crowded areas. Third, shorter papers consistently showed higher LLM-modification compared to longer papers, which may indicate that researchers working under time constraints are more likely to rely on AI for writing assistance. These results may be an indicator of the competitive nature of certain research areas and the pressure to publish quickly.

   If the majority of modification comes from an LLM owned by a private company, there could be risks to the security and independence of scientific practice. 
  We hope our results inspire 
further studies of widespread LLM-modified text and conversations about how to promote transparent, epistemically diverse, accurate, and independent scientific publishing.

\paragraph{Limitations} While our study focused on ChatGPT, which accounts for more than three-quarters of worldwide internet traffic in the category \citep{vanrossum2024generative}, we acknowledge that there are other large language models used for assisting academic writing. 
Furthermore, while \cite{Liang2023GPTDA} demonstrate that GPT-detection methods can falsely identify the writing of language learners as LLM-generated, our results showed that consistently low false positives estimates of $\alpha$ in 2022, which contains a significant fraction of texts written by multilingual scholars. We recognize that significant author population changes~\citep{Globalaitalent} or other language-use shifts could still impact the accuracy of our estimates. Finally, the associations that we observe between LLM usage and paper characteristics are correlations which could be affected by other factors such as research topics. More causal studies is an important direction for future work.

\subsection*{Acknowledgments}
We thank Daniel A. McFarland, Dan Jurafsky, Zachary Izzo, Xi Victoria Lin, Lingjiao Chen, and Haotian Ye for their helpful comments and discussions. J.Z. is supported by the National Science Foundation (CCF 1763191 and CAREER 1942926), the US National Institutes of Health (P30AG059307 and U01MH098953) and grants from the Silicon Valley Foundation and the Chan-Zuckerberg Initiative. and H.L. is supported by the National Science Foundation (2244804 and 2022435) and the Stanford Institute for Human-Centered Artificial Intelligence (HAI).
\clearpage

\bibliography{colm2024_conference}
\bibliographystyle{colm2024_conference}

\appendix

\clearpage
\newpage

\clearpage

\section{Estimated Fraction of LLM-Modified Sentences in \textit{Introductions}}

\begin{figure}[ht!] 
    \centering
    \includegraphics[width=1.00\textwidth]{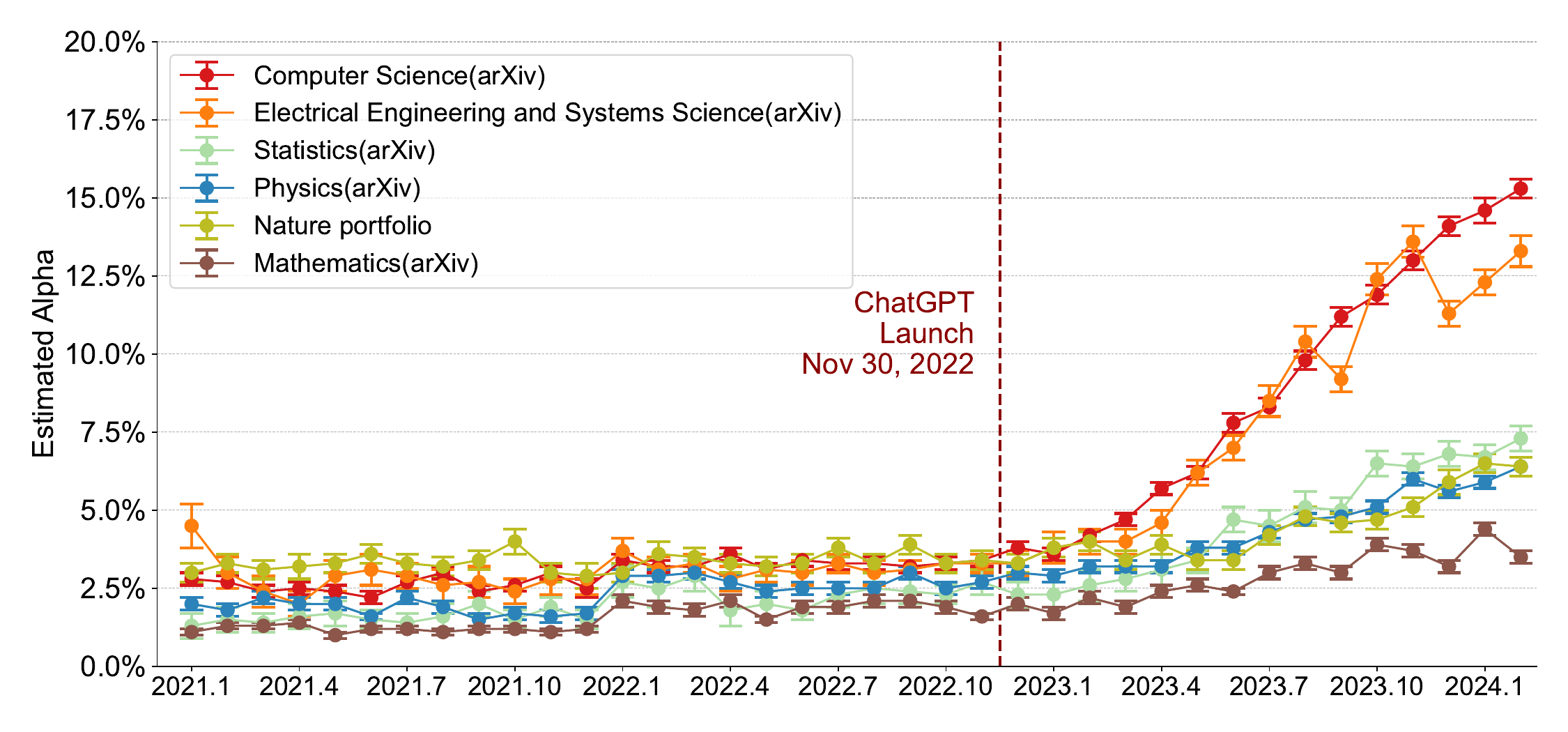}
    \caption{
    \textbf{Estimated Fraction of LLM-Modified Sentences in \textit{Introductions} Across Academic Writing Venues Over Time. }
    We focused on the introduction sections for the main texts, as the introduction was the most consistently and commonly occurring section across different categories of papers.
    This figure presents the estimated fraction ($\a$) of sentences in introductions which are LLM-modified, across the same venues as Figure~\ref{fig: temporal-abstract}. 
    We found that the results are consistent with those observed in abstracts (Figure~\ref{fig: temporal-abstract}). 
    We did not include \textit{bioRxiv} introductions as there is no bulk download of PDFs available.
    Error bars indicate 95\% confidence intervals by bootstrap.
    }
    \label{fig: temporal-introduction}
\end{figure}

\clearpage

\section{LLM prompts used in the study}

\begin{figure}[htb!]
\begin{lstlisting}
The aim here is to reverse-engineer the author's writing process by taking a piece of text from a paper and compressing it into a more concise form. This process simulates how an author might distill their thoughts and key points into a structured, yet not overly condensed form. 

Now as a first step, first summarize the goal of the text, e.g., is it introduction, or method, results? and then given a complete piece of text from a paper, reverse-engineer it into a list of bullet points.
\end{lstlisting}
\caption{
Example prompt for summarizing a paragraph from a human-authored paper into a skeleton: This process simulates how an author might first only write the main ideas and core information into a concise outline. The goal is to capture the essence of the paragraph in a structured and succinct manner, serving as a foundation for the previous prompt.
}
\label{fig:skeleton-prompt-1}
\end{figure}

\begin{figure}[htb!]
\begin{lstlisting}
Following the initial step of reverse-engineering the author's writing process by compressing a text segment from a paper, you now enter the second phase. Here, your objective is to expand upon the concise version previously crafted. This stage simulates how an author elaborates on the distilled thoughts and key points, enriching them into a detailed, structured narrative. 

Given the concise output from the previous step, your task is to develop it into a fully fleshed-out text.
\end{lstlisting}
\caption{
Example prompt for expanding the skeleton into a full text: The aim here is to simulate the process of using the structured outline as a basis to generate comprehensive and coherent text. This step mirrors the way an author might flesh out the outline into detailed paragraphs, effectively transforming the condensed ideas into a fully articulated section of a paper. The format and depth of the expansion can vary, reflecting the diverse styles and requirements of different academic publications.
}
\label{fig:skeleton-prompt-2}
\end{figure}

\begin{figure}[htb!]
\begin{lstlisting}
Your task is to proofread the provided sentence for grammatical accuracy. Ensure that the corrections introduce minimal distortion to the original content. 
\end{lstlisting}
\caption{
Example prompt for proofreading.
}
\label{fig:proofread-prompt}
\end{figure}

\clearpage

\section{Additional Information on Implementation and Validations}
\label{appendix:sec:implementation}

\paragraph{Supplementary Information about Data}
We collected data for this study from three publicly accessible sources: official APIs provided by \textit{arXiv} and \textit{bioRxiv}, and web pages from the \textit{Nature} portfolio. 
For each of the five major \textit{arXiv} categories (Computer Science, Electrical Engineering and Systems Science, Mathematics, Physics, Statistics), we randomly sampled 2,000 papers per month from January 2020 to February 2024. 
Similarly, from \textit{bioRxiv}, we randomly sampled 2,000 papers for each month within the same timeframe. 
For the \textit{Nature} portfolio, encompassing 15 \textit{Nature} journals including Nature, Nature Biomedical Engineering, Nature Human Behaviour, and Nature Communications, we followed the same sampling strategy, selecting 2,000 papers randomly from each month, from January 2020 to February 2024. 
The procedure for generating the AI corpus data for a given time period is described in aforementioned Section $\S$~\ref{main:sec: method}.

When there were not enough papers to reach our target of 2,000 per month, we included all available papers. 
The \textit{Nature} portfolio encompasses the following 15 \textit{Nature} journals: 
\textit{Nature, 
Nature Communications, 
Nature Ecology \& Evolution, 
Nature Structural \& Molecular Biology, 
Nature Cell Biology, 
Nature Human Behaviour, 
Nature Immunology, 
Nature Microbiology, 
Nature Biomedical Engineering, 
Communications Earth \& Environment, 
Communications Biology, 
Communications Physics, 
Communications Chemistry,
Communications Materials,}
and \textit{Communications Medicine}.

\paragraph{Additional Information on Large Language Models}

In this study, we utilized the gpt-3.5-turbo-0125 model, which was trained on data up to September 2021, to generate the training data for our analysis. The LLM was employed solely for the purpose of creating the training dataset and was not used in any other aspect of the study.

We chose to focus on ChatGPT due to its dominant position in the generative AI market. According to a comprehensive analysis conducted by FlexOS in early 2024, ChatGPT accounts for an overwhelming 76\% of global internet traffic in the category, followed by Bing AI at 16\%, Bard at 7\%, and Claude at 1\% \citep{vanrossum2024generative}. This market share underscores ChatGPT's widespread adoption and makes it a highly relevant subject for our investigation. Furthermore, recent studies have also shown that ChatGPT demonstrates substantially better understanding of scientific papers than other LLMs~\citep{liang2023can,liu2023reviewergpt}.

We chose to use GPT-3.5 for generating the training data due to its free availability, which lowers the barrier to entry for users and thereby captures a wider range of potential LLM usage patterns. This accessibility makes our study more representative of the broad phenomenon of LLM-assisted writing. 
Furthermore, the previous work by \citet{liang2024monitoring} has demonstrated the framework's robustness and generalizability to other LLMs. Their findings suggest that the framework can effectively handle significant content shifts and temporal distribution shifts.

Regarding the parameter settings for the LLM, we set the decoding temperature to 1.0 and the maximum decoding length to 2048 tokens during our experiments. The Top P hyperparameter, which controls the cumulative probability threshold for token selection, was set to 1.0. Both the frequency penalty and presence penalty, which can be used to discourage the repetition of previously generated tokens, were set to 0.0. Additionally, we did not configure any specific stop sequences during the decoding process.

\newpage 
\clearpage

\section{Word Frequency Shift in arXiv Computer Science introductions}
\begin{figure}[ht!]
    \centering
    \includegraphics[width=1.00\textwidth]{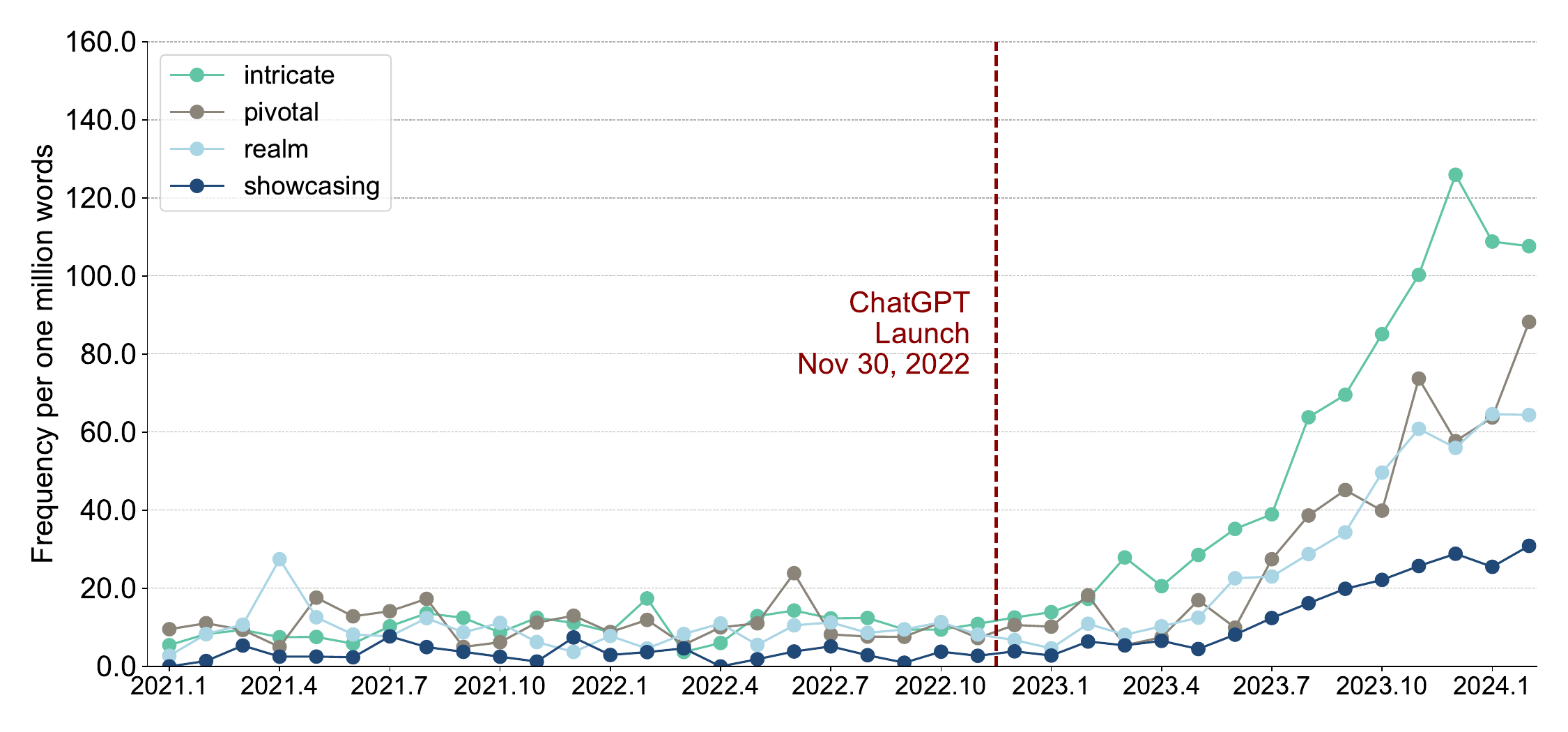}
\caption{
    \textbf{Word Frequency Shift in sampled \textit{arXiv} Computer Science introductions in the past two years.} 
The plot shows the frequency over time for the same 4 words as demonstrated in Figure \ref{fig: arxiv-revisions}. The words are: \textit{realm}, \textit{intricate}, \textit{showcasing}, \textit{pivotal}. 
The trend is similar for two figures.
Data from 2010-2020 is not included in this analysis due to the computational complexity of parsing the full text from a large number of arXiv papers.
}
\end{figure}

\clearpage
\newpage

\section{Fine-grained Main Findings}

\begin{figure}[htb!] 
    \centering
    \includegraphics[width=1.00\textwidth]{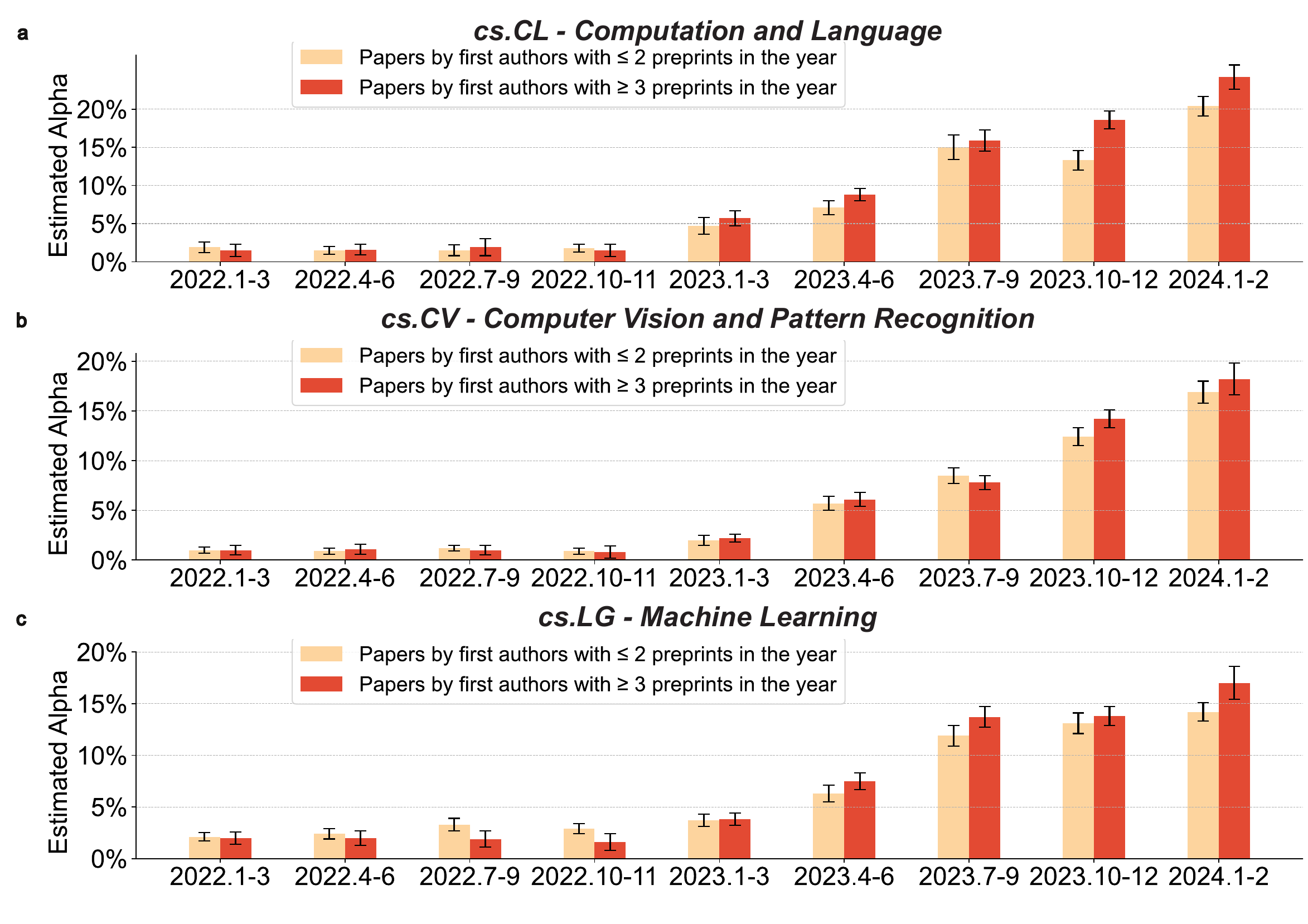}
    \caption{    
\textbf{The relationship between first-author preprint posting frequency and LLM usage holds across \textit{arXiv} Computer Science sub-categories.}
Papers in each \textit{arXiv} Computer Science sub-category (cs.CV, cs.LG, and cs.CL) are stratified into two groups based on the preprint posting frequency of their first author, as measured by the number of first-authored preprints in the year: those with $\leq 2$ preprints and those with $\geq 3$ preprints. 
Error bars indicate 95\% confidence intervals by bootstrap.
    }
    \label{supp:figure:upload}
\end{figure}

\begin{figure}[htb!] 
    \centering
    \includegraphics[width=1.00\textwidth]{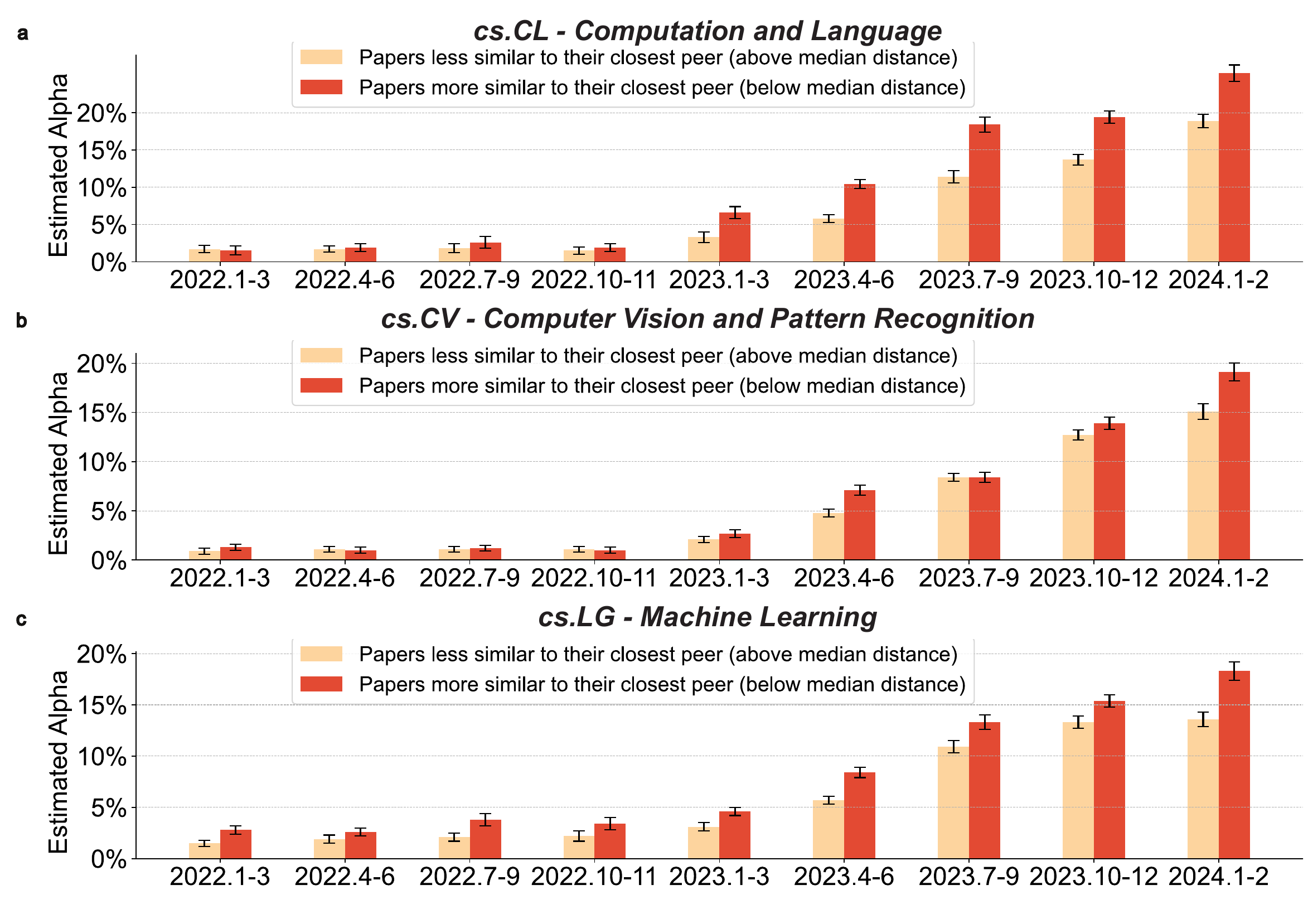}
    \caption{    
\textbf{The relationship between paper similarity and LLM usage holds across \textit{arXiv} Computer Science sub-categories.}
Papers in each \textit{arXiv} Computer Science sub-category (cs.CV, cs.LG, and cs.CL) are divided into two groups based on their abstract's embedding distance to their closest peer within the respective sub-category: papers more similar to their closest peer (below median distance) and papers less similar to their closest peer (above median distance). 
Error bars indicate 95\% confidence intervals by bootstrap.
    }
    \label{supp:figure:crowded}
\end{figure}

\begin{figure}[htb!] 
    \centering
    \includegraphics[width=1.00\textwidth]{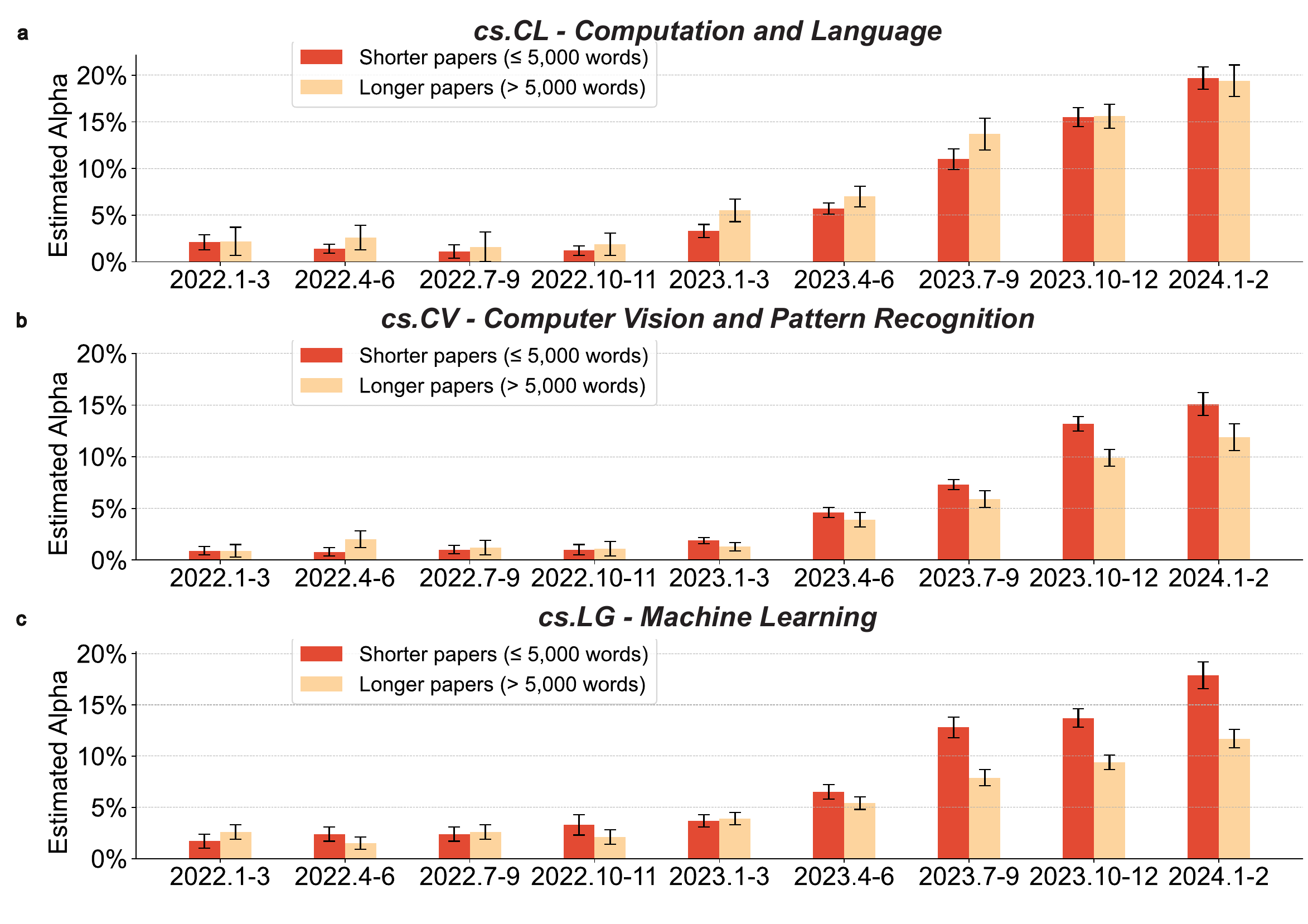}
    \caption{    
\textbf{The relationship between paper length and LLM usage holds for cs.CV and cs.LG, but not for cs.CL.}
Papers in each \textit{arXiv} Computer Science sub-category (cs.CV, cs.LG, and cs.CL) are stratified by their full text word count, including appendices, into two bins: below or above 5,000 words (the rounded median). 
For cs.CL, no significant difference in LLM usage was found between shorter and longer papers, possibly due to the limited sample size, as only a subset of the PDFs were parsed to calculate the full length.
Error bars indicate 95\% confidence intervals by bootstrap.
    }
    \label{supp:figure:length}
\end{figure}
\newpage 
\clearpage

\section{Proofreading Results on arXiv data}
\label{supp:sec:proofreading}
\begin{figure}[ht!]
    \centering
    \includegraphics[width=1.00\textwidth]{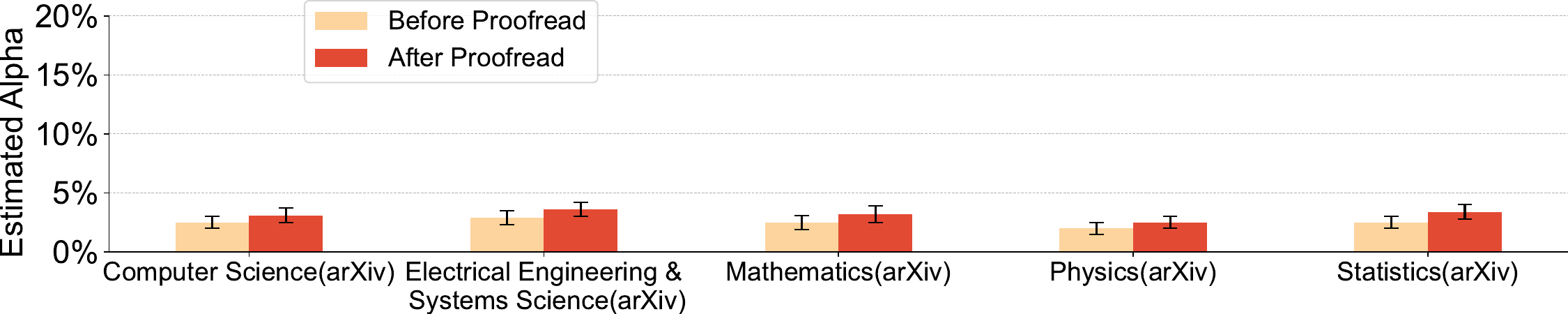}
\caption{
    \textbf{Robustness of estimations to proofreading.} 
The plot demonstrates a slight increase in the fraction of LLM-modified content after using Large Language Models (LLMs) for ``proofreading" across different \textit{arXiv} main categories. This observation validates our method's robustness to minor LLM-generated text edits, such as those introduced by simple proofreading. 
The analysis was conducted on 1,000 abstracts from each \textit{arXiv} main category, randomly sampled from the period between January 1, 2022, and November 29, 2022.
Error bars indicate 95\% confidence intervals by bootstrap.
}
\label{supp:figure:proofreading}
\end{figure}

\newpage 
\clearpage

\section{Extended Related Work}
\label{appendix:sec:related-work}

\paragraph{Zero-shot LLM detection.} A major category of LLM text detection uses statistical signatures that are characteristic of machine-generated text, and the scope is to detect the text within individual documents. Initially, techniques to distinguish AI-modified text from human-written text employed various metrics, such as entropy \citep{Lavergne2008DetectingFC}, the frequency of rare n-grams \citep{Badaskar2008IdentifyingRO}, perplexity \citep{Beresneva2016ComputerGeneratedTD}, and log-probability scores \citep{solaiman2019release}, which are derived from language models. More recently, DetectGPT \citep{Mitchell2023DetectGPTZM} found that AI-modified text is likely to be found in areas with negative log probability curvature. DNA-GPT \citep{Yang2023DNAGPTDN} improves performance by examining the divergence in n-gram patterns. Fast-DetectGPT \citep{Bao2023FastDetectGPTEZ} enhances efficiency by utilizing conditional probability curvature over raw probability. \citet{Tulchinskii2023IntrinsicDE} studied the intrinsic dimensionality of generated text to perform the detection. We refer to recent surveys by \citet{Yang2023ASO, Ghosal2023TowardsP} for additional details and more related works. However, zero-shot detection requires direct access to LLM internals to enable effective detection. Closed-source commercial LLMs, like GPT-4, necessitate using proxy LLMs, which compromises the robustness of zero-shot detection methods across various scenarios \citep{Sadasivan2023CanAT, Shi2023RedTL, Yang2023ASO, Zhang2023AssayingOT}.

\paragraph{Training-based LLM detection.} Another category is training-based detection, which involves training classification models on datasets that consist of both human and AI-modified texts for the binary classification task of detection. Early efforts applied classification algorithms to identify AI text across various domains, such as peer review submissions \citep{Bhagat2013SquibsWI}, media publications \citep{Zellers2019DefendingAN}, and other contexts \citep{Bakhtin2019RealOF, Uchendu2020AuthorshipAF}. Recently, researchers have finetuned pretrained language model backbones for this binary classification. GPT-Sentinel \citep{Chen2023GPTSentinelDH} uses the constructed dataset OpenGPTText to train RoBERTa \citep{Liu2019RoBERTaAR} and T5 \citep{raffel2020exploring} classifiers. GPT-Pat \citep{Yu2023GPTPT} trains a Siamese neural network to compute the semantic similarity of AI text and human text. \citet{Li2023DeepfakeTD} build a wild testbed by gathering texts from various human writings and texts generated by different LLMs. Using techniques such as contrastive and adversarial learning can enhance classifier robustness \citep{Liu2022CoCoCM, Bhattacharjee2023ConDACD, Hu2023RADARRA}. We refer to recent surveys \citet{Yang2023ASO, Ghosal2023TowardsP} for additional methods and details. However, these publicly available tools for detecting AI-modified content have sparked a debate about their effectiveness and reliability~\citep{OpenAIGPT2,jawahar2020automatic,fagni2021tweepfake,ippolito2019automatic,mitchell2023detectgpt,human-hard-to-detect-generated-text,mit-technology-review-how-to-spot-ai-generated-text,survey-2023, solaiman2019release}. OpenAI's decision to discontinue its AI-modified text classifier in 2023 due to ``low rate of accuracy'' further highlighted this discussion~\citep{Kirchner2023,Kelly2023}.

Training-based detection methods face challenges such as overfitting to training data and language models, making them vulnerable to adversarial attacks \citep{Wolff2020AttackingNT} and biased against non-dominant language varieties \citep{Liang2023GPTDA}. The theoretical possibility of achieving accurate \textit{instance}-level detection has also been questioned~\citep{Weber-Wulff2023,Sadasivan2023CanAT,chakraborty2023possibilities}.

\paragraph{LLM watermarking.} 
Text watermarking introduces a method to detect AI-modified text by embedding an imperceptible signal, known as a watermark, directly into the text. This watermark can be retrieved by a detector that shares the model owner's secret key. Early watermarking techniques included synonym substitution \citep{Chiang2003NaturalLW, Topkara2006TheHV} and syntactic restructuring \citep{Atallah2001NaturalLW, Topkara2006NaturalLW}. Modern watermarking strategies involve integrating watermarks into the decoding process of language models \citep{aaronson, kirchenbauer2023watermark, Zhao2023Ginsew}. Researchers have developed various techniques, such as the Gumbel watermark \citep{aaronson}, which uses traceable pseudo-random softmax sampling, and the red-green list approach \citep{kirchenbauer2023watermark, Zhao2023ProvableRW}, which splits the vocabulary based on hash values of previous n-grams. Some methods focus on preserving the original token probability distributions \citep{ Hu2023UnbiasedWF,Kuditipudi2023RobustDW, Wu2023DiPmarkAS}, while others aim to improve detectability and perplexity \citep{zhao2024permute} or incorporate multi-bit watermarks \citep{Yoo2023RobustMN, Fernandez2023ThreeBT}. However, one major concern with watermarking is that it requires the involvement of the model or service owner, such as OpenAI, to implant the watermark during the text generation process. 
In contrast, the framework by \cite{liang2024monitoring} operates independently of the model or service owner's intervention, allowing for the monitoring of AI-modified content without requiring their active participation or adoption.

\paragraph{Implications for LLM Pretraining Data Quality} 

The increasing prevalence of AI-modified content in academic papers, particularly on platforms like \textit{arXiv}, has important implications for the quality of LLM pretraining data. \textit{arXiv} has become a significant source of training data for LLMs, contributing approximately 2.5\% of the data for models like Llama \citep{touvron2023llama}, 12\% for RedPajama \citep{elazar2023s}, and 8.96\% for the Pile \citep{gao2020pile}. Our findings suggest that a growing proportion of this pretraining data may contain LLM-modified content. Preliminary research indicates that the inclusion of LLM-modified content \citep{veselovsky2023artificial} in LLM training can lead to several pitfalls, such as the reinforcement of stereotypes and biases against anyone who is not a middle-aged ``European/North American man" \citep{ghosh2023person, santurkar2023whose}, the flattening of variation in language and content \citep{dell2023navigating}, and the potential failure of models to accurately capture the true distribution of the original content, which may result in model collapse \citep{shumailov2023curse}. \cite{santurkar2023whose} demonstrate that this phenomenon amplifies the effect of LLMs providing content that is unrepresentative of most of the world. As such, our results underscore the 
importance of robust data curation and filtering strategies even in seemingly unpolluted datasets.

\end{document}